\documentclass[letterpaper, 10 pt, conference]{ieeeconf}  

\IEEEoverridecommandlockouts                              

\usepackage{graphicx} 
\graphicspath{{images/}}
\usepackage{amsmath} 

\title{\LARGE \bf
ONE PIECE: One Patchwork In Effectively Combined Extraction for grasp
}

\author{Xiao Hu$^{1,2}$,HangJie Mo$^{\dag}$,XiangSheng Chen$^{1,3}$,JinLiang Chen$^{\dag}$,and Xiangyu Chen$^{\dag}$ 
}

\begin{document}

\maketitle
\thispagestyle{empty}
\pagestyle{empty}

\begin{abstract}

For grasp network algorithms, generating grasp datasets for a large number of 3D objects is a crucial task. However, generating grasp datasets for hundreds of objects can be very slow and consume a lot of storage resources, which hinders algorithm iteration and promotion. For point cloud grasp network algorithms, the network input is essentially the internal point cloud of the grasp area that intersects with the object in the gripper coordinate system. Due to the existence of a large number of completely consistent gripper area point clouds based on the gripper coordinate system in the grasp dataset generated for hundreds of objects, it is possible to remove the consistent gripper area point clouds from many objects and assemble them into a single object to generate the grasp dataset, thus replacing the enormous workload of generating grasp datasets for hundreds of objects. We propose a new approach to map the repetitive features of a large number of objects onto a finite set.To this end, we propose a method for extracting the gripper area point cloud that intersects with the object from the simulator and design a gripper feature filter to remove the shape-repeated gripper space area point clouds, and then assemble them into a single object. The experimental results show that the time required to generate the new object grasp dataset is greatly reduced compared to generating the grasp dataset for hundreds of objects, and it performs well in real machine grasping experiments. We will release the data and tools after the paper is accepted.

\end{abstract}

\section{INTRODUCTION}
For humans, \cite{c1} grasping is an intuitive and simple task, but for robots, grasping is not easy.\cite{c2}\cite{c3} However, for both humans and robots, what matters is the graspable region on the target object.\cite{c4}Only when the graspable region is available, can the robot execute the grasping task. There are many grasping algorithms that focus on obtaining better graspable regions and performing grasping on them. \cite{c5}\cite{c7}\cite{c8}For grasping network algorithms, generating grasp datasets is crucial, and it directly determines the quality of the algorithm. However, unlike 2D images, generating grasp datasets has always been a very resource-consuming task.

There are mainly two ways to generate grasp datasets for 3D objects, which are manual annotation and data-driven methods. Both methods are based on the external, interactive shape of the grasping object. Manual annotation has\begin{figure}[h]
        \centering
        \includegraphics[width=0.48\textwidth]{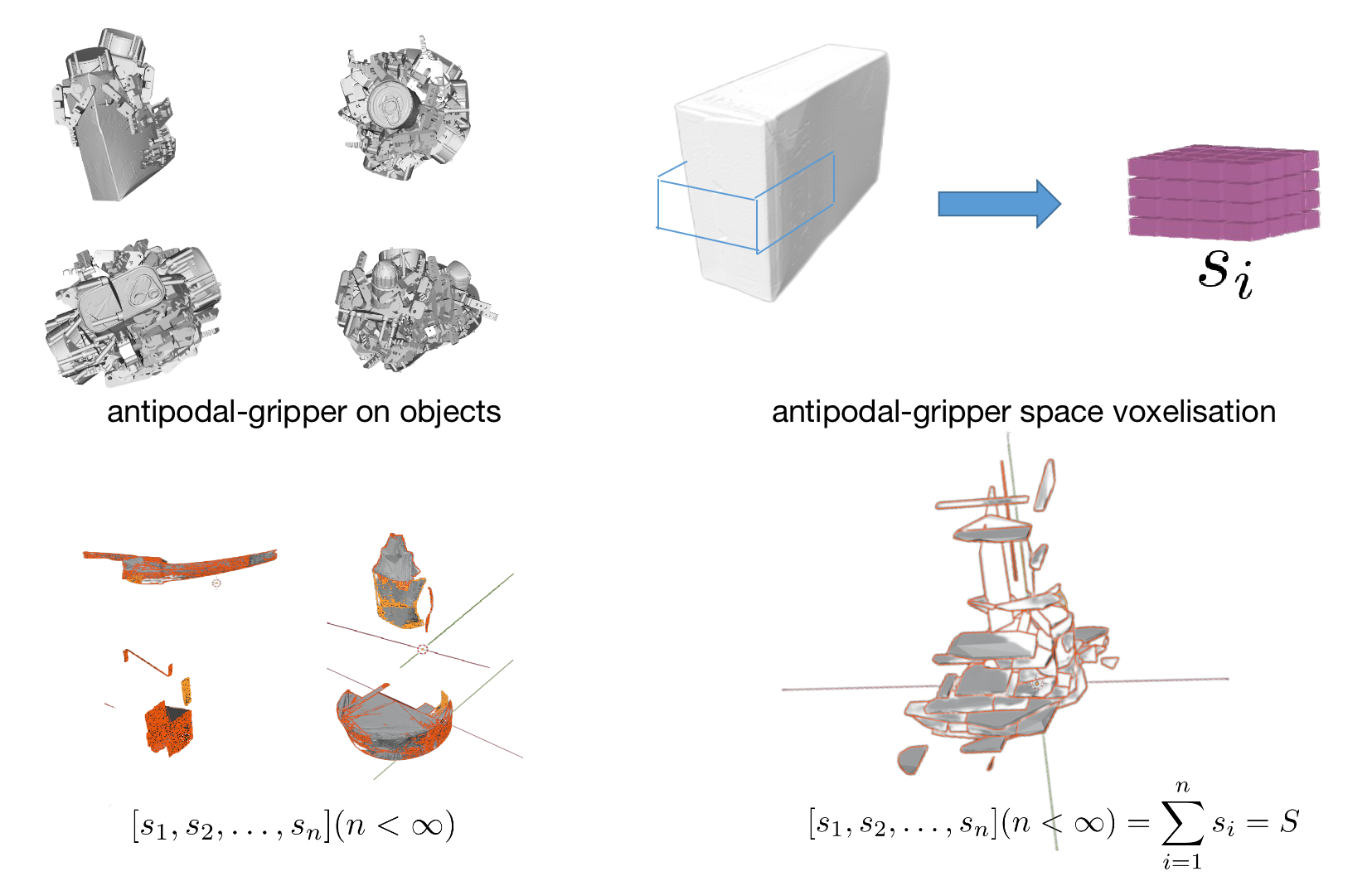} 
        \caption{{\bf Overview of process}:Since the jaw-based point cloud is finitely distributed in the jaw coordinate system, duplicate jaw point clouds are sieved out and assembled into an object based on this.} 
        \label{fig:example} 
\end{figure}
the characteristics of prior knowledge guidance and strong subjectivity, but the resource consumption is very high. Annotating grasps for 3D objects requires a lot of thinking and judgment, which makes the consumption of human labor huge. The other method is data-driven. This method relies on constructing grasp points and vectors on the surface mesh of objects to generate grasp datasets for 3D objects, \cite{c9}and uses the method of form closure and force closure to score grasp configurations. The problem is that the grasp configurations generated may not be all real and effective due to the contact between the gripper and the object during grasping. More importantly, the data-driven method requires the traversal of each grasp point for calculation, which makes the generation of grasp datasets very slow. It often takes several days to generate grasp datasets for dozens of objects, which is very disadvantageous for the iteration and use of grasping algorithms.

To date, for grasping network algorithms,\cite{c10} the intersection of claw point clouds with objects in the claw coordinate system and corresponding annotations or scores are used as the basis for constructing the dataset, and the network model is constructed to perform regression and classification on the point clouds that intersect with objects in the claw coordinate system. Since in the claw coordinate system, all possible point cloud shapes inside the voxelized claw space at the required resolution for the corresponding task form a finite element set, for multiple objects, the repeated shapes correspond to an element of this finite set. The redundant shapes cause redundancy, so the finite element set in the claw coordinate system can be extracted and assembled into an object to replace the previous multiple objects, \cite{c11}greatly reducing storage space and decreasing computational resource consumption.

In this paper, we propose a complete process to obtain stable grasping regions in the grasp coordinate system of 3D objects \cite{c12}\cite{c13}and remove redundant grasp point clouds from the grasp and object intersection. Finally, these grasp point clouds are assembled into an object to replace the generation of grasping datasets for many objects and improve the generation speed. The specific process is as follows: first, we propose an Isaac Gym-based framework to obtain stable grasping regions and design a gripper-feature-area feature extractor to filter out identical point clouds based on the grasp coordinate system, extract non-redundant point clouds of the intersection between the gripper and the object, and assemble them into an object.

Our main contributions are:

\begin{itemize}
\item[$\bullet$]We designed a method based on simulator to extract stable grasping regions that are different from those obtained from static grasping configurations.
\item[$\bullet$] We proposed a novel gripper-feature-area extractor that can filter out similar grasping shape features.
\item[$\bullet$]The generated single object can significantly improve the speed of generating grasping datasets, greatly reduce the storage requirements, and achieve a relative good grasping success rate in real-world grasping experiments.
\end{itemize}

\section{RELATED WORK}
Over the years, grasp has been a popular research topic in robotics. Most of the research has been focused on generating datasets for grasping based on SE(2) and SE(3) for robotic arms. The core of these datasets is providing grasp annotations for a large number of objects with different shapes.
\subsection{Grasp Datasets in SE(2)}
The Cornell Dataset \cite{c18}consists of 885 RGB-D images of 240 different objects, with 8019 hand-labeled grasp rectangles.The dataset can be used to train neural networks to detect grasps in images. However, compared to traditional datasets used in deep learning, the dataset with only 885 images is very small and may lead to poor performance in generalizing to different images or object configurations. The hand-labeled grasps may also be biased towards grasps that are easy for humans to execute, but not necessarily using parallel-jaw grippers.

The Jacquard Dataset\cite{c19} can be considered a larger and more objective version of the Cornell Dataset. It contains 54K images generated in simulation using a mixer, using 11K unique object mesh models. Each image is labeled with multiple grasp rectangles that roughly cover the space of possible grasps (using the same representation as in Cornell), for a total of 110M grasp labels. The labels are generated by evaluating the quality of a large number of possible grasps, sampled randomly and simulated in PyBullet. Like Cornell, the performance of Jacquard is measured as accuracy at 25\% or 30\% IOU detection thresholds.

To overcome the time-consuming data generation problem, Mahler et al. created Dexnet-2.0\cite{c26}, a synthetic dataset of 6.7 million depth images labeled with grasps that succeeded when executed at the center of the image. These images were created using 1500 mesh models generated in simulation. Each depth image is labeled with up to 100 grasp detections obtained using grasp planning methods on the underlying mesh model. These object models were selected from the 3DNet and KIT object databases and extracted from 50 different object categories. The dataset is augmented with Gaussian noise at each pixel and with 180-degree image reflections and rotations.

Due to the complexity of the factors involved, such as the object, the scene, and the grasp pose, datasets of this type are difficult to cover all cases, leading to limitations. Therefore, it is necessary to combine other methods for dataset generation and algorithm learning and validation.

\subsection{Object datasets for SE(3) grasping}

Early research on grasp prediction assumed that robots have a perfect understanding of their environment and aimed to plan grasps based on the 3D models of objects \cite{c28}\cite{c29}. Using this technique, Goldfeder et al. \cite{c27} created the Columbia Grasp Database, which contains over 230k grasps. These methods only used the CAD models of the gripper and the object. This may be the earliest grasp dataset for SE(3) grasping \cite{c32}. It consists of 7256 object mesh models extracted from the Princeton Shape Benchmark \cite{c33}. A total of 238k SE(3) grasps were marked on the 7256 models using the GRAPIT! simulator\cite{c34}.

YCB (Yale, Columbia, Berkeley)\cite{c16} is an object set rather than a grasp dataset, but it is used so frequently that we include it here. YCB is a set of 77 household objects for which 3D mesh models and RGB images are available online.
The Berkeley Dex-Net project was published with the following three object sets : 1) 1.3k synthetic 3D mesh models from a 50 category subset of 3DNet; 2) 129 mesh models from the KIT Object Database; 3) 13 "adversarial" models.

EGAD! is an interesting dataset comprised of 2331 objects \cite{c20}. The focus here is on an algorithmic approach to generating 3D object geometries that span a space of shape complexity and grasp difficulty. In particular, each object is ranked from 1 to 25 according to both measures (shape complexity and grasp difficulty) and placed on a 25 × 25 grid. The 2331 objects were generated while ensuring that the dataset contains no more than four objects belonging to any single grid square.

GraspNet-1Billion is another grasp dataset worth mentioning \cite{c24}. Although it includes data drawn from only 88 objects (32 of which are from YCB and another 13 from DexNet 2.0), it is paired with 97k RGBD images of 190 different cluttered scenes containing the objects. Each of these RGBD images is labeled with tens of thousands of SE(3) grasp poses, for a total of 1.1B labeled grasp examples.

ACRONYM is a recent SE(3) dataset comprised of 17.7 million labeled grasps performed on 8872 object meshes\cite{c23}.\begin{figure*}[htbp]
        \center{\includegraphics[width=16cm]  {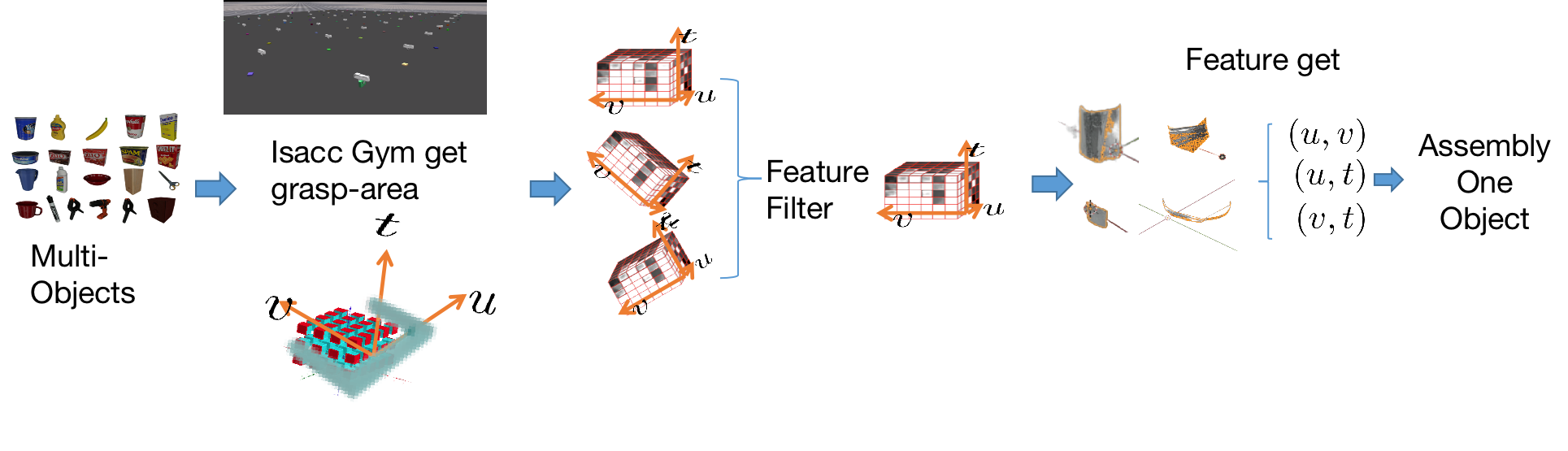}} 
        \caption{{\bf Our Framework}:We extract features from multiple objects, filter out duplicate features using a feature selector to obtain a finite set of object point clouds in the gripper coordinate system, and then assemble them using a classifier.}
        \label{fig}
\end{figure*}  
The object meshes were obtained from ShapeNet\cite{c21}. The grasps were labeled using the FleX physics simulator\cite{c11}to simulate the Franka Panda gripper (maximum grip aperture of 8cm). Observations are available either as depth images or point clouds, produced using PyRender2. A notable aspect of this dataset is that it does not include data in the standard dense clutter setting.

Object-based grasp dataset generation methods can generate large amounts of annotated data and cover various object shapes and surface features, thereby improving the performance of robot grasping tasks. However, this method also has some limitations, such as the potential distribution differences between synthetic and real data, which may reduce the algorithm's generalization ability. Therefore, this method needs to be combined with real data for training and validation to improve the algorithm's performance.

\section{PROBLEM Description}

Point cloud grasping algorithms are a hot research topic in the field of robotics. With the development of artificial intelligence technologies, such as deep learning, many advanced point cloud grasping algorithms have been proposed and widely applied. However, the training of the network component in these point cloud grasping algorithms requires a significant amount of storage and computational resources.

Taking PointNetGPD \cite{c14}as an example, this algorithm performs feature extraction on the point cloud and uses a neural network-based pose regression module to predict the grasping pose and position of the object. Finally, it outputs the grasping configuration for execution. PointNetGPD mainly consists of the following parts: GRASP Candidates Generation, PointNet feature extractor, and output. In PointNetGPD, the PointNet network structure is used, which can directly operate on point cloud data, avoiding the need for data conversion.

In the entire process, the most time-consuming step is the first step of GRASP Candidates Generation. In this step, a large number of grasping configurations and corresponding grasping point clouds are generated. Each grasping point cloud is scored based on its shape closure and force closure, and this process requires a significant amount of time for computation. Therefore, we hope to simplify this process.\begin{figure}[h]
        \centering
        \includegraphics[width=0.2\textwidth]{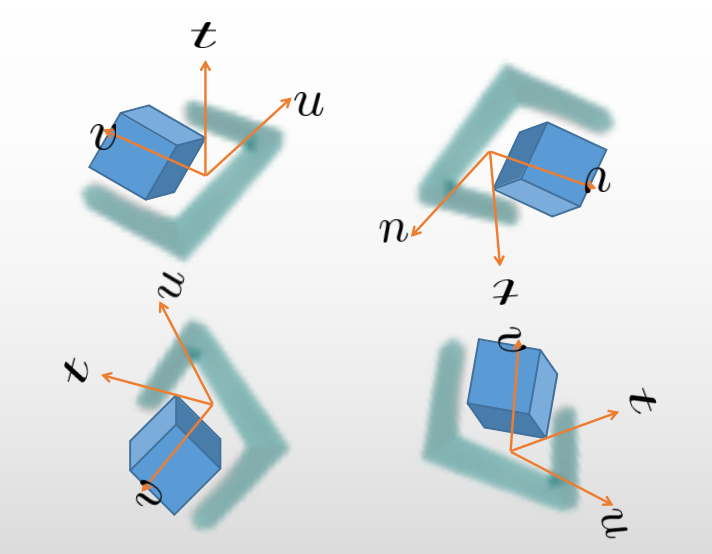} 
        \caption{Spatial coordinates of the jaws based on parallel jaws do not change with parallel jaw posture} 
        \label{fig:example} 
\end{figure}

Since the grasping point clouds extracted are represented in the gripper coordinate system, i.e.,$(u,v,t)$, and all are based on the gripper coordinate system, the coordinates of the grasping point cloud set $p_{i}\in(u,v,t)$ under the corresponding grasping configuration are unchanged when the gripper grasping configuration rotates and translates in the base coordinate system$(x,y,z)$. Therefore, many $p_{i}$ are consistent and repetitive. Based on this characteristic, we designed a feature extractor to filter out identical $p_{i}$. For the remaining $p_{i}$, sampling and scoring are performed and then trained, which is equivalent to obtaining a weakly supervised classifier.\cite{c28} Although some generalization ability is sacrificed, significant improvements are obtained in storage and data set generation time. This facilitates the use of grasping network algorithms because most people want to quickly deploy and use the network algorithm without wasting a significant amount of time and resources on data downloading, data set generation, and storage.

Based on the above reasons, we hope to optimize the repeated elements in a large collection of objects, extract all non-repeating point clouds in the gripper coordinate system on a large number of objects, and assemble them into one object.

\begin{equation}
    argmin[s_1, s_2, \ldots, s_n] (n < \infty)
\end{equation}

\section{ONE SINGLE OBJECT GENERATION}

\subsection{Object Supplement}
\begin{figure}[h]
        \centering
        \includegraphics[width=0.4\textwidth]{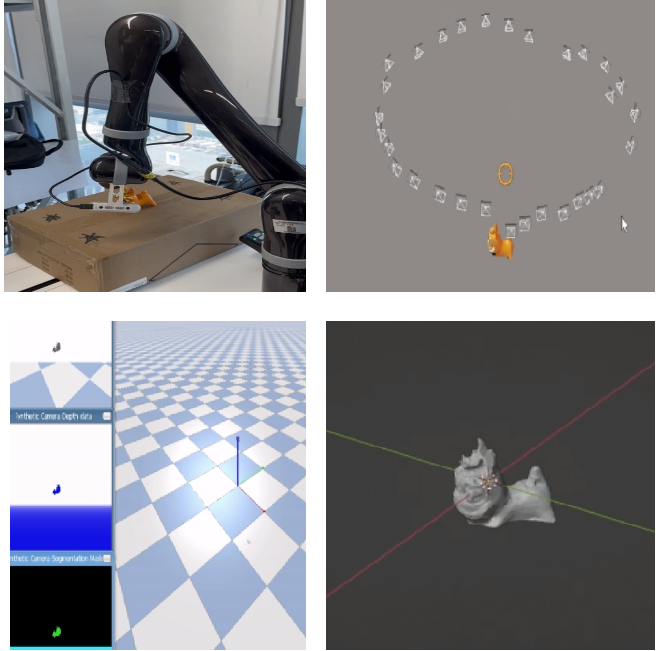} 
        \caption{{\bf Overview of Object Supplement}:Using robotic arms for fully automated object capture, CAD model generation, and related data collection.} 
        \label{fig:example} 
\end{figure}

Currently, most publicly available datasets online consist of everyday objects, while some specific tasks may require objects with highly distinctive features that are completely different from everyday objects. This can result in the loss of important features for downstream tasks, which may significantly reduce the effectiveness of algorithms that were supposed to perform well. Therefore, it is necessary to supplement the existing object datasets according to downstream tasks.

In addition, for our proposed method, due to the large size of CAD datasets and the relatively poor maintenance of open-source datasets, some everyday objects with significant grasping features are missing in the downloading process.\cite{c29} Therefore, we use this method to supplement the missing everyday objects with significant grasping features to ensure the completeness and applicability of the CAD model dataset. Then we process the complete CAD model dataset to extract and generate object models.

We provide a fast and convenient method that can generate CAD models and depth images for objects that need to be supplemented. The specific process is as follows: first, the mechanical arm is used to densely capture images of the object based on depth information. Second, the Nerf method is used to process the captured images and generate CAD models. Finally, the CAD models are imported into pybullet\cite{c30} to generate depth images at different angles.

This supplementary process can generate CAD models that comply with file formats and usage requirements, and effectively supplement objects with highly distinctive features that are missing in publicly available CAD datasets. This method can be applied to various specific tasks, providing more comprehensive data support for research and practice in robot grasping algorithms.

\subsection{Extraction}
\begin{figure}[h]
        \centering
        \includegraphics[width=0.48\textwidth]{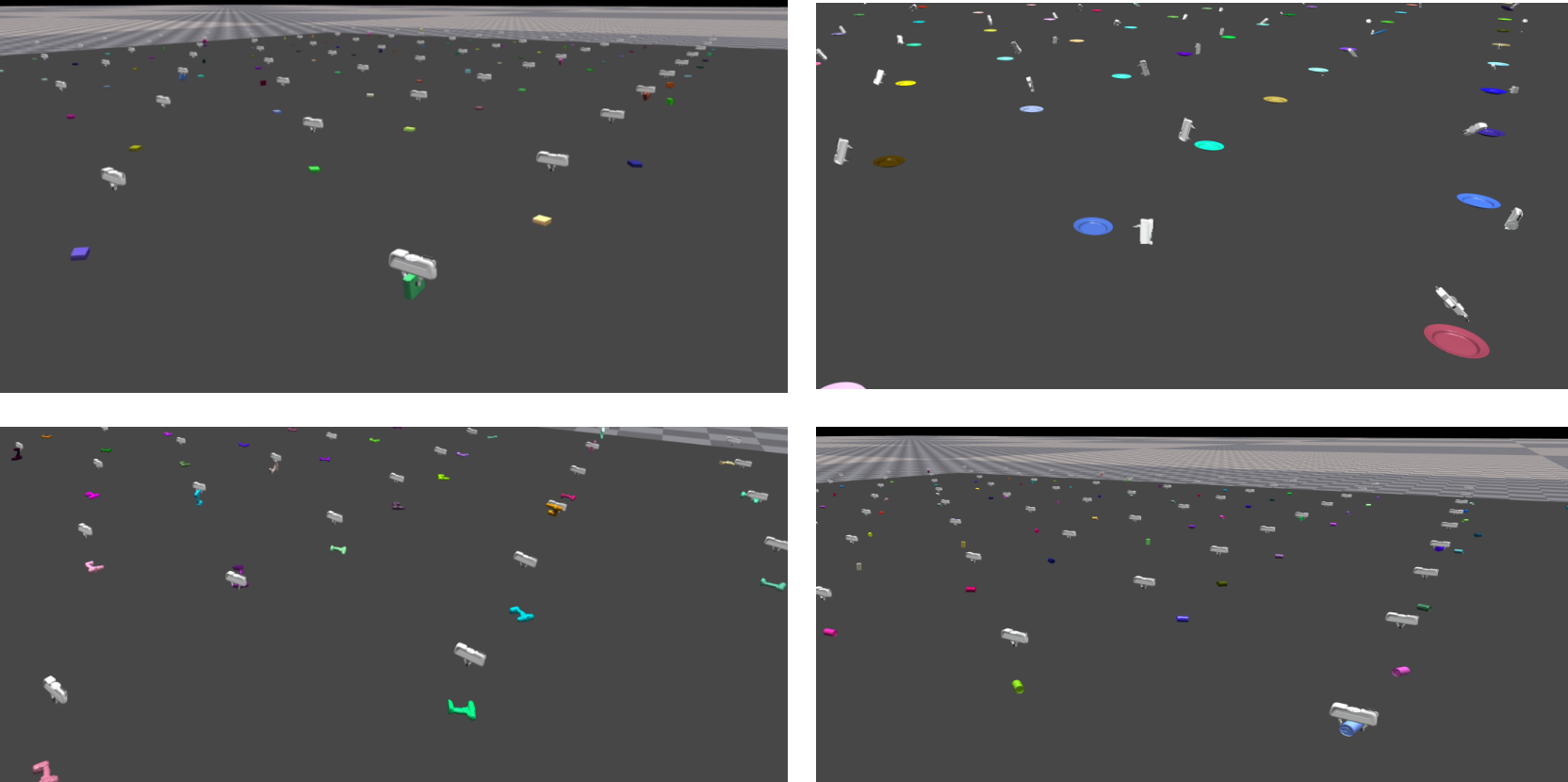} 
        \caption{Extraction of the area where the object intersects the jaws from the object by random gripping in Issac Gym\cite{c31}} 
        \label{fig:example} 
\end{figure}
Using the advantages of graphics simulators and GPU computing, we randomly sample grasping points for each object and generate grasp configurations based on the normal vectors constructed by nearby points. These large amounts of sampled grasp configurations are simulated in Isaac Gym to extract the point cloud of gripper intersections with objects. Compared to traditional methods such as GPD and GPG, this method has the advantages of clear process and faster speed.

In addition, we extracted the part of the CAD model dataset that can interact with the gripper and removed the gripper point cloud that cannot intersect with the point cloud of the object. For these parts that cannot intersect with the object, $p=\sum_{i}^{n}(x,y,z)(n=0)$, since they cannot form a dataset, they also waste computing resources.

For CAD models generated under static conditions, the grasping area under dynamic conditions is different from that under static conditions, $[R,w]p_{(u,v,t)} \ne p_{(u,v,t)}$. When the gripper is closed, it will come into contact with the object, causing a small displacement of the object and thus a change in the grasping area. Therefore, it is necessary to obtain the grasping area where the gripper intersects with the object. Under static grasping configurations, the grasping area where the gripper intersects with the object may be an invalid area, because when this grasping area is used for grasping, the obtained grasping area is completely different from the static grasping configuration. Moreover, the scoring or labeling of the generated grasping dataset is also based on the static grasping configuration, which may result in a large number of errors. Ignoring the correspondence between dynamic grasping configuration and the grasping area where the gripper intersects with the object will introduce noise into the grasping network, because the grasping network assumes that there is a one-to-one correspondence between the grasping configuration and the gripper point cloud. With this method, we can obtain the noise under dynamic conditions and perform denoising processing on the subsequent generated grasping dataset. The method we designed ensures that the extracted gripper point cloud is in the grasping and intersecting area of the object and will not give high scores or good labels to the wrong areas during scoring or labeling.

\subsection{Grippers-Feature-Area Filters}
In the gripper coordinate system, there are many gripper regions that intersect with the object and have the same shape. These gripper regions with identical shapes are redundant in the input dataset of the network and therefore need to be removed. To solve this problem, we construct a feature filter to remove the overlapping gripper regions that are the same in the object's coordinate system at a specified resolution of voxel space.

Firstly, the gripper is voxelized in the gripper space, and the voxelization is performed according to the downsampling scale corresponding to the resolution of the grasping network algorithm. Here, we take the example of cm scale to illustrate the feature filter. First, the voxelization expression of the gripper space is constructed:$Length_{gripper}=a,Width_{gripper}=b,Height_{gripper}=c$.

Therefore, the total number of voxels in the gripper space $Sum_{voxel}=a*b*c$

Based on the voxel space at the corresponding resolution, encoding is performed. If a point falls within the voxel grid, the voxel is represented as 1, and if there is no point within the voxel, it is represented as 0. Therefore, the encoded voxel space can be represented as:

\begin{figure}[h]
        \centering
        \includegraphics[width=0.3\textwidth]{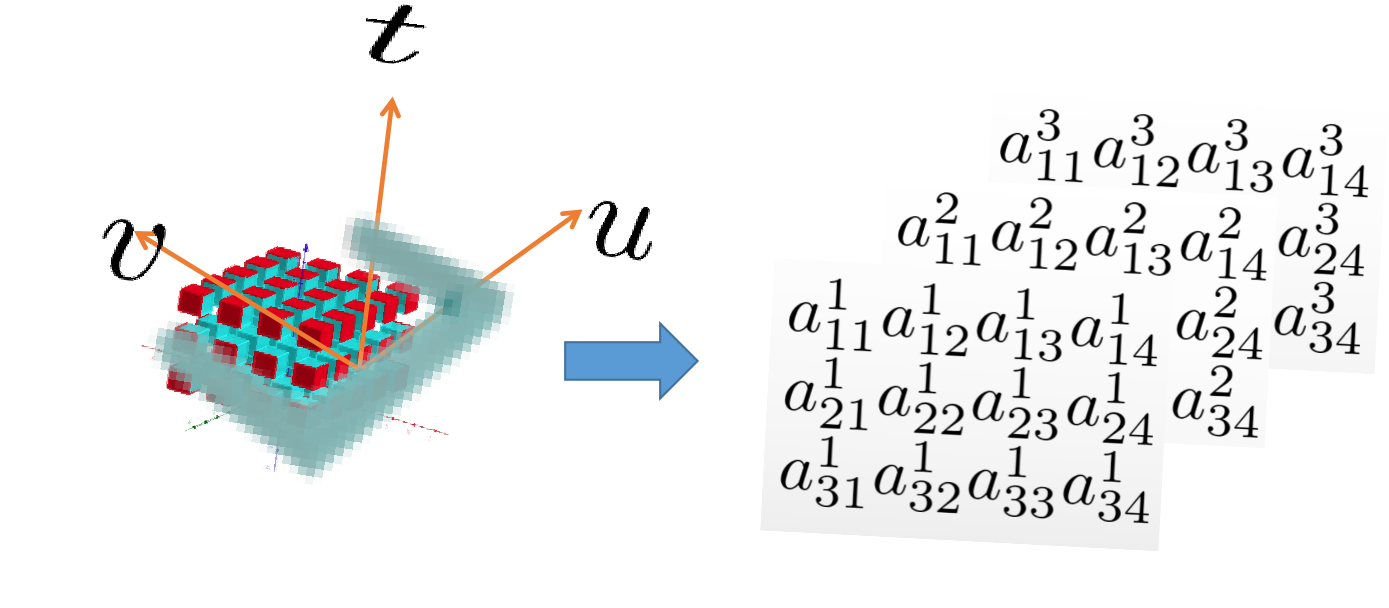} 
        \caption{Gripper Space code} 
        \label{fig:example} 
\end{figure}

Then, the following constraints are used to construct the feature selector to determine whether $A,a_{ij}\in [0,1]$ and $A_n^{*},a_{nij}\in [0,1]$ are consistent for the encoded grasp point clouds in the claw space.

\begin{equation}
    \begin{cases}
        A_{n}={\displaystyle J}-A_{n}^{*}\\
        a_{ij}^{a_{n_ij}}=Sum-Sum_{A}\\
        a_{ij}+a_{nij}=Sum\\
    \end{cases}
\end{equation}

We define a matrix power operation between two matrices $a_{1}$ and $a_{n}$, which performs element-wise power and then sum. Additionally, we reverse the values of one of the matrices and substitute the resulting matrix into the constraint, where the dimensionality of the resulting matrix is the same as that of the unit matrix subtracted from $a_{n}$. If the constraint is satisfied, the two matrices are identical, indicating that the shape of the grasping space is the same, and thus the regions of the object intersecting with the gripper that have the same shape are filtered out.

\subsection{Assemble One Object}
\begin{figure}[h]
        \centering
        \includegraphics[width=0.3\textwidth]{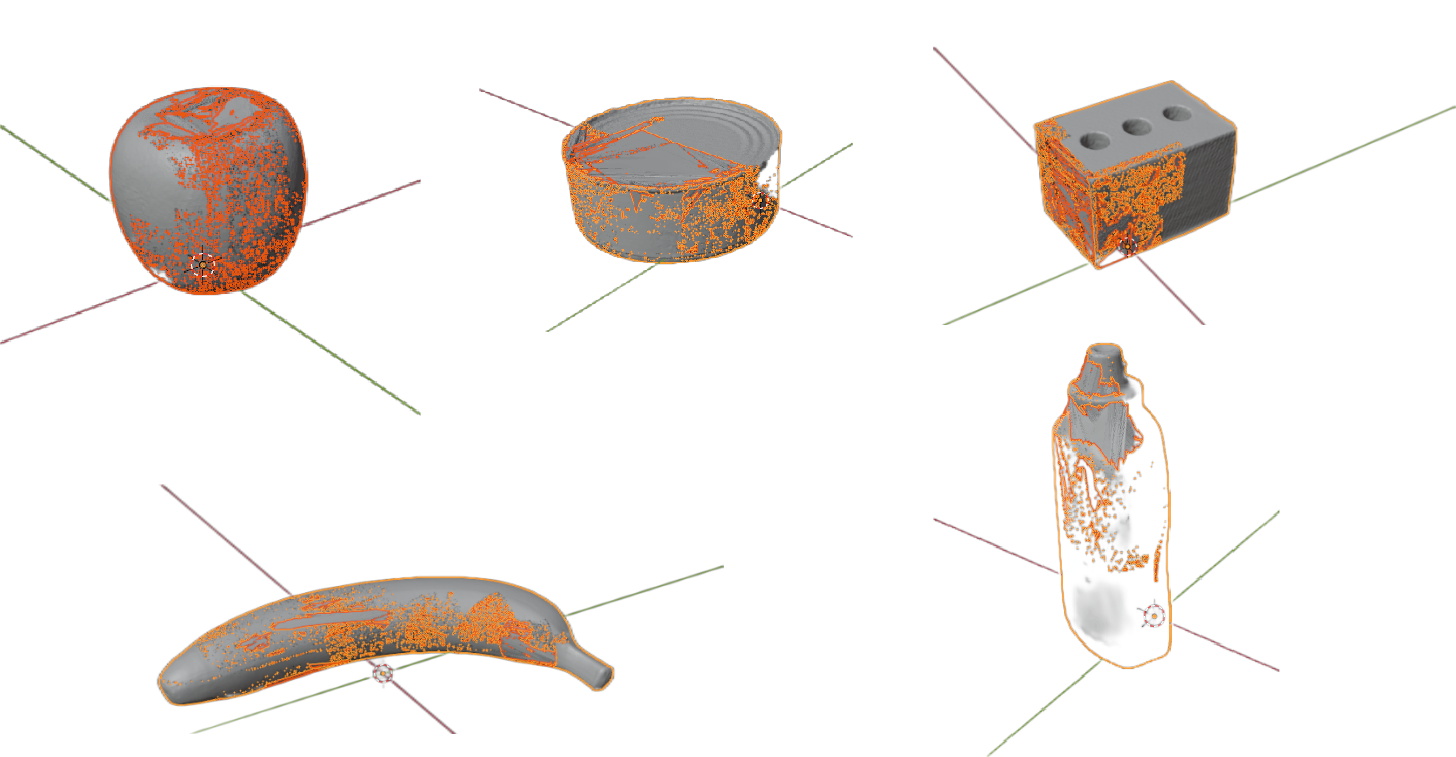} 
        \caption{{\bf Gripper Space} The highlighted area on the object is the area corresponding to the jaws that intersect the object.} 
        \label{fig:example} 
\end{figure}

After extracting all the grasping regions that intersect with objects using the simulator Isaac Gym and filtering out regions with redundant shapes using the above method, the next step is to assemble them into an object. The main problem to consider when assembling is how to handle duplicate feature positions in space, as otherwise many features will be obscured by stacking.

\begin{figure}[h]
        \centering
        \includegraphics[width=0.2\textwidth]{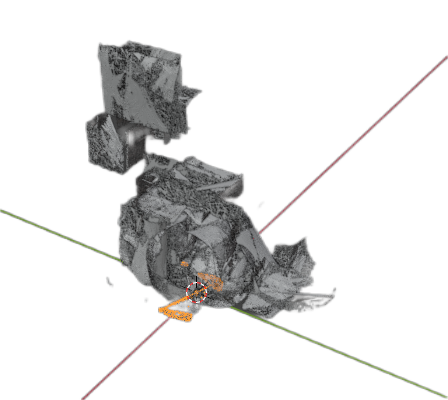} 
        \caption{Stacking of feature areas.stacking result in a large number of features being obscured} 
        \label{fig:example} 
\end{figure}

To this , we designed a set of assembly classifiers that classify the projected features on the three gripper planes, $(u,v), (u,t), (v,t)$ as the most important features in the gripper space $(u,v,t)$. During the grasping process, the projection of the gripper point cloud on the three different planes is not the same in the gripper coordinate system, $P_{uv}\neq P_{ut}\neq P_{vt}$. Based on this, each gripper point cloud that intersects with the object is projected onto these three planes in the gripper coordinate system, and then features are extracted for classification. After classification, based on the richness of different projection planes, better assembly can be achieved, avoiding the problem of feature masking.

The classifier is designed to extract the most favorable features for grasp generation at the corresponding resolution. For example, if a grasp feature region $P_{i}$ has the most distinct features on a certain projection plane, then this grasp feature region is assembled on that plane to effectively avoid the problem of feature vectors stacking and masking. We use voxelized gripper space with a convolution kernel of $weight=1$ to extract features on the $(v,t)$ plane.

We use the classifier to perform convolutional sliding on the voxelized point cloud projections. When the sum of the convolutional results is equal to 3, we increment the $N_{feature}$ counter. The classifier uses the total number of features to classify the grasping point cloud, obtaining the required edge features, and then assembles the features on the projection plane with the most edge features.

\begin{equation}
    \begin{cases}
       \begin{bmatrix}
        1 & 1 \\
        1 & 0
        \end{bmatrix}
        \begin{bmatrix}
        1 & 1 \\
        0 & 1
        \end{bmatrix}
        \begin{bmatrix}
        1 & 0 \\
        1 & 1
        \end{bmatrix}
        \begin{bmatrix}
        0 & 1 \\
        1 & 1
        \end{bmatrix}
    \end{cases}
\end{equation}

\begin{figure*}[htbp]
        \center{\includegraphics[width=16cm]  {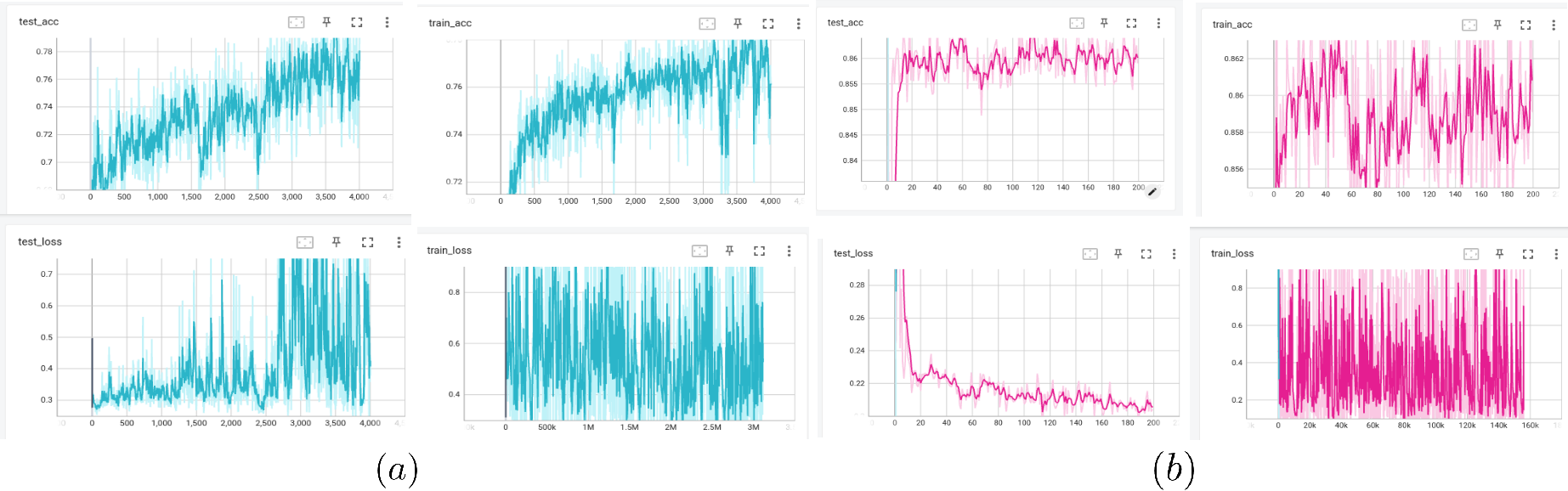}} 
        \caption{Training effects using the same test set,{\bf(a)}:64 objects.{\bf(b)}:ours.Our generated objects show good convergence performance based on training results. In order to accelerate the learning speed of the network, we set the learning rate to a relatively large value, which led to the divergence phenomenon. However, after iterating many times, we were able to achieve model convergence.}
        \label{fig}
\end{figure*}

To assemble each feature, we compared the areas of each projection $S_{u,v}$, $S_{u,t}$, and $S_{v,t}$ in the grasping space coordinate system. Specifically, we compared the ratio $S/(S_{u,v}+S_{u,t}+S_{v,t})\times N_{feature}$ and selected the projection area with the maximum ratio as the assembly plane for the corresponding feature. This ensures that the grasping features are maximally preserved. We repeated this process for all features.

\section{EXPERIMENT}
To validate the effectiveness of the object, we designed an experiment to compare the performance of the object generated using this method with 64 other objects, including 61 objects from the open source YCB dataset, supplemented with three objects using the method described earlier: an elliptical glasses case, a doll, and a milk carton. The method was used to generate and assemble these 64 objects into a single object, and the performance was compared with that of two other CAD object datasets.

\begin{table}[h]
\caption{COMPARISON}
\label{table_example}
\begin{center}
\begin{tabular}{ccc}
Objects & Generation time & Storage\\
\hline
Ours & About Six hours & 6.9GB\\
\hline
64 Objects & About Two Days & 43.7GB\\
\end{tabular}
\end{center}
\end{table}

\begin{figure}[h]
        \centering
        \includegraphics[width=0.48\textwidth]{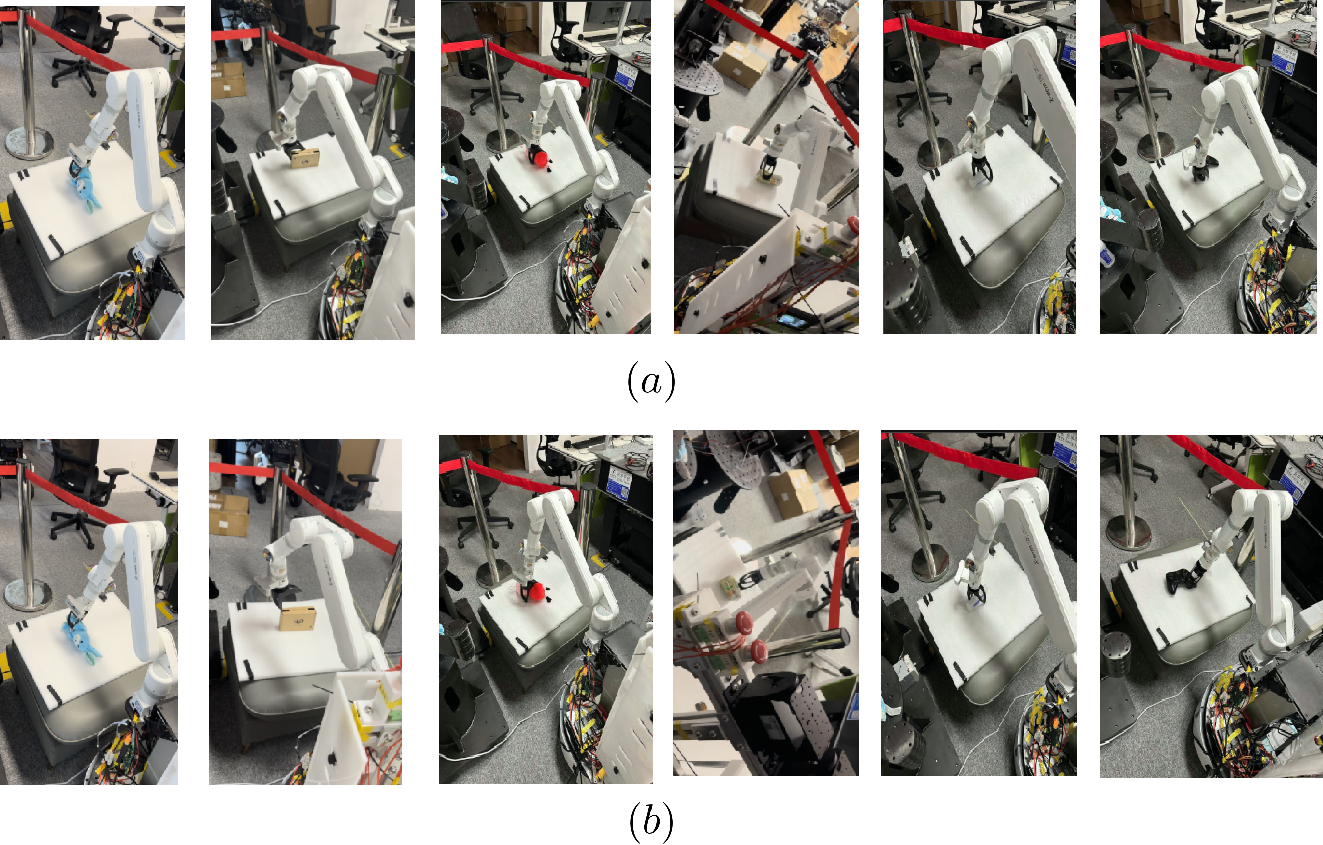} 
        \caption{{\bf Grasp Experiments}:{\bf(a)}:64 objects.{\bf(b)}:ours.Grasping tests on novel and unseen objects} 
        \label{fig:example} 
\end{figure}

During the actual grasping experiments, the shape of the grasping space and the intersection with the object in our trained model is consistent with the features extracted from the model directly trained on all objects. For example, for the box, the grasping positions are all at the corner positions, while for the doll, the grasping positions are all at the center position of the doll. This validates that the performance of the PointNet network\cite{c22}\cite{c23} in classifying our generated objects is consistent with that of directly using a large number of objects trained, but our objects have a much better storage and speed for generating grasping datasets, with certain sacrifice of generalization performance that has little impact on actual grasping results, greatly reducing the consumption of resources, demonstrating the significance of this work.

We conducted real-world tests on 19 objects, with two trials for each successful grasp and three trials for each failed grasp. The grasping performance was consistent, and both models exhibited consistent reasons for failed grasps.The results of the experiment are as follows.

\begin{table}[h]
\caption{RESULTS OF GRASPING EXPERIMENTS}
\label{table_example}
\begin{center}
\begin{tabular}{ccc}
Models & Success rate & Completion rate\\
\hline
Ours & 89\% & 94\%\\
\hline
64 Objects & 89\% & 94\%\\
\end{tabular}
\end{center}
\end{table}

In the experiment, the grasp configuration of the failed grasp attempts was consistent. From the experimental results, the single object assembled by our proposed method showed the same performance in real-world experiments as models trained on multiple objects.

\section{Discussion And Future Work}

We propose a novel approach to map the repetitive features of a large number of objects onto a finite set. Although we have only partially demonstrated the applicability of this method from one perspective, we believe that this approach has enormous potential for exploration, and future work will test this method on different types of datasets and evaluate the generated objects on different grasping network algorithms. \cite{c9}We believe that this work has great scalability because the storage of 3D objects has always been a big problem, whether for personal computers or the cloud. Therefore, this method can significantly improve the iteration and popularization of grasping algorithms.

In the experimental section, due to unstable experimental equipment and time constraints, we did not perform enough experiments, but we found through experimental observations that the position of each grasp of our generated objects was highly consistent with that of the 64 objects. \cite{c16}\cite{c19}This reflects that the features of the extracted objects are highly aggregated with those of the 64 objects in shape. In the future, we will continue to improve the experimental design and increase the number of experiments to further verify the proposed method.

In addition, unlike images, 3D objects may produce two completely different photos due to shooting angles, sensor errors, and other reasons. However, the features and overall characteristics of 3D objects will not change due to external factors. Therefore, mapping methods can be designed to reduce the required computational resources when using 3D objects.

This work also has certain inspirations for current popular generative models. Many features of 3D objects may be repetitive, so these non-repetitive parts can be used to construct a finite set for a specified downstream task. By combining this finite set with a generative model, the desired object can be directly generated, while reducing resource consumption in storage and other aspects.

\addtolength{\textheight}{-12cm}   




\end{document}